\documentclass{article} 
\usepackage{nips15submit_e,times}
\usepackage{hyperref}
\usepackage{url}
\usepackage{subcaption}
\usepackage{epsfig}
\usepackage{amssymb}
\usepackage{multirow}

\title{INsight: A Neuromorphic Computing System for Evaluation of Large Neural Networks}

\author{
Jaeyong Chung
\\
Department of Electronic Engineering\\
Incheon National University\\
Incheon, Korea 406-772 \\
\texttt{jychung@inu.ac.kr} \\
\And
Taehwan Shin \\
Department of Electronic Engineering\\
Incheon National University\\
Incheon, Korea 406-772 \\
\AND
Yongshin Kang \\
Department of Electronic Engineering\\
Incheon National University\\
Incheon, Korea 406-772 \\
}

\nipsfinalcopy 
\newcommand{\RNum}[1]{\uppercase\expandafter{\romannumeral #1\relax}}

\begin{document}

\maketitle
\begin{abstract}
Deep neural networks have been demonstrated impressive results in various cognitive tasks such as object detection and image classification. In order to execute large networks, Von Neumann computers store the large number of weight parameters in external memories, and processing elements are timed-shared, which leads to power-hungry I/O operations and processing bottlenecks.
This paper describes a neuromorphic computing system that is designed from the ground up for the energy-efficient evaluation of large-scale neural networks. The computing system consists of a non-conventional compiler, a neuromorphic architecture, and a space-efficient microarchitecture that leverages existing integrated circuit design methodologies. The compiler factorizes a trained, feedforward network into a sparsely connected network, compresses the weights linearly, and generates a time delay neural network reducing the number of connections. The connections and units in the simplified network are mapped to silicon synapses and neurons. We demonstrate an implementation of the neuromorphic computing system based on a field-programmable gate array that performs the MNIST hand-written digit classification with 97.64\% accuracy.
  
\end{abstract}

\section{Introduction}

Large-scale neural networks such as deep convolutional neural networks have been shown state-of-the-art results on various tasks
 in  
computer vision, automatic speech recognition, natural language processing, and audio recognition, and on certain tasks, their performance has become comparable to humans~\cite{ciresan2012multi}. However, these networks contain a huge number of parameters (e.g., $10^8$~\cite{sermanet2013overfeat}), and the evaluation of such models is computationally expensive. This makes it problematic to deploy these models to embedded platforms, where computing power, memory, storage, and energy are limited. It is also problematic to evaluate the large models at the server side. For example, processing images and videos uploaded by millions of users require massive amounts of computation, and running a data center that supports such computation carries enormous costs including cooling expenses and  electricity bills. 

The evaluation of large neural networks  can be accelerated using general-purpose graphical processing units (GPGPU), and several custom accelerators have been proposed as well~\cite{farabet2010hardware, chakradhar2010dynamically}. However, all these processing units are based on Von Neumann architecture, and the large amount of parameters are stored in high-density external memories. The traditional architecture requires inefficient data transfer between parameter-storing memories and processing units, leading to high-energy consumption and processing bottleneck~\cite{backus1978can}.

As neural networks are models inspired by brains, it is natural to execute them in the processing units that mimic brains. Neuromorphic architectures emulate biological neural networks, and they locate memory and processing elements in a close proximity. This type of architectures has potential for eliminating the issues in the today's computers~\cite{merolla2014million}. However, mixed analog-digital neuromorphic systems such as BrainScaleS~\cite{schemmel2010wafer} and Neurogrid~\cite{benjamin2014neurogrid} are not designed to run real-world applications, but aim to simulate the brain for study on computational neuroscience. They usually adopt biologically plausible neuron models. The neuron models at a low-level of abstractions are considered expressive and versatile as biological neurons do, but it comes at an implementation cost such as complexity, space, and energy. To run specific applications, certain details are not necessary.  TrueNorth~\cite{arthur2012building, esser2013cognitive, amir2013cognitive, andrew2013cognitive}, the neuromorphic architecture from IBM, aims to balance the computational capability and the cost judiciously, and adopts digital spiking neurons based on the leaky integrate-and-fire model. A recently unveiled chip~\cite{merolla2014million} based on TrueNorth contains 4096 custom-designed cores and employs asynchronous circuits for low-power operations. The chip is used for both the brain simulation and cognitive applications~\cite{cassidy2014real}.

This paper presents a novel neuromorphic computing system that is designed solely for the execution (i.e., inference or testing) of deep neural networks.  To the best of our knowledge, this is the first neuromorphic computing system that stores all the weight parameters on-chip by compressing neural networks. Designing a new computer architecture is a complex task and has many trade-offs. Our design philosophy to deal with it is summarized in the next section. 

\section{Design Considerations}

The performance of neural networks such as classification accuracy is extremely valuable. As will be shown later, in some way, even 1\% degradation of the performance is translated into a huge saving of the implementation cost, and we do not want to sacrifice the performance in other ways.  The performance of learning algorithms for spiking neurons such as spike timing dependent plasticity (STDP)~\cite{seo201145nm} is not close to the state of the art yet~\cite{brader2007learning}. While non-spiking neural networks trained using state-of-the-art algorithms can be converted into spiking neural networks, constraints arisen from the hardware architecture and inefficiency in information coding could degrade the performance~\cite{arthur2012building}.  

We desire to build an ``artificial'' ``brain''. How much should we learn from brains and what should we throw away from the architecture of existing processors?  This may be the central question in designing neuromorphic architectures. We take a moderate approach to this problem.  As much as we want to learn from the nature, we also desire to take advantage of the results of humans' effort over several decades. Since we still make the artificial brain into a silicon chip, we need to continue to use techniques optimized for structures of silicon chips.  Until today, the man-made information processing device at the lowest power consumption is application-specific integrated circuits (ASICs) and, for general purpose, field programmable gate arrays (FPGAs). Thus, we leverage the ASIC/FPGA design methodology, and balance the ASIC/FPGA architecture and the structure of brains judiciously.

Our neuromorphic computing system aims at the evaluation (i.e., the forward-execution) of neural network models and is not intended to train the models. Training a neural network is computationally more expensive than the forward-execution, and there have been research efforts to accelerate the training using FPGAs~\cite{ly2009high, kim2014fully}. Our main objective is low-power signal processing rather than high-performance computing, although it can provide high performance in power-limited environments including data centers. For various applications such as object recognition, a model can be trained using a cluster of CPU or GPU-based machines, and copies of the model can provide desired functionality without modification. We also do not consider on-line learning. STDP facilitates efficient on-line learning, but as above-mentioned, it does not provide the state-of-the-art performance.

Our neuromorphic system is digital. Digital systems use a high level of abstraction on signals, which allows us to deal with complexity and makes it easier to ensure the correctness of man-made systems. While analog silicon neurons can provide a higher integration density than that of digital neurons, we make it digital for those benefits.

 The rest of the paper is organized as follows. Section~3 describes the motivation of our work. Section~4,~5,~and~6 propose the compiler, the neuromorphic architecture, and the microarchitecture of our computing system, respectively. Section~7 shows experimental results, and Section~8 concludes the paper.
 
\section{Motivation}
Each unit (node) in feedforward neural networks mainly performs weighted sum operations, and the functionality can be easily implemented in a hardware module using logic multipliers, logic adders, and weight-storing registers. For each unit in the model, we create an instance of the module and these instances are interconnected following the model, yielding a circuit design. However, this type of systems has not built until today for the following reasons. 1) Neural networks for interesting applications are huge. Recent deep neural network architectures contain millions and billions of parameters. Each parameter at least requires a logic multiplier and a register to store the parameter, and we cannot integrate such a high number of multipliers and registers in a chip. For this reason, existing accelerators of neural networks store parameters in high-density external memories and equip with a small number of processing elements. The external memory access draws much power, and the limited memory I/O bandwidth becomes a processing bottleneck. Also, the processing elements are time-shared, which necessitates complex scheduling to maximize the utilization and creates another processing bottleneck. 2) Neural networks are densely connected unlike logic (circuit) networks.  Densely interconnected designs cause routing congestion issues in the conventional ASIC/FPGA design methodology. For this reason, existing neuromorphic architectures~\cite{snider2008molecular} rely on custom-designed crossbars, where inputs and outputs are fully connected. 3) Silicon chips have limited I/O bandwidth and all the inputs of neural networks are usually not available at the same time. For example, traditional cameras transfer each pixel at a time, but neural network models for processing images take all the pixels of a whole image as input in parallel. Thus, the direct realization of neural network models results in systems with very low utilization.
\begin{figure}
\centering
\epsfig{figure = 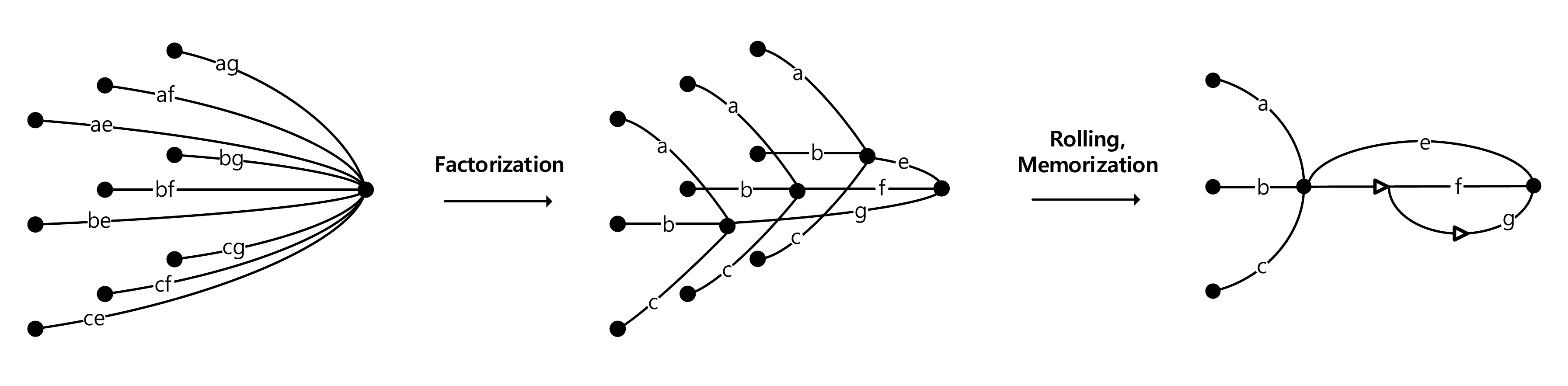, width = 5.5in}
\caption{A fully connected feedforward network of a single layer is transformed into a sparsely connected network with two layers by factorization. The number of parameters is reduced from 9 to 6 by factorization, but the number of connections rather increases from 9 to 12. However, when it is converted into a TDNN, the number of connections with parameters becomes 6. The reduction of the connections is critical in neuromorphic architectures, where each connection uses a dedicated computational resource called the synapse.
}\label{ex1}
\end{figure}

Fortunately, weight parameters in neural networks are heavily redundant~\cite{denil2013predicting}, i.e., these networks are highly over-parameterized. This is necessary to obtain a high quality model during training by providing much flexibility to learning algorithms, but for the forward-execution, such a high number of parameters are not necessary. Our neuromorphic architecture is mainly motivated by the fact that the high redundancy in weight parameters can be exploited to minimize the implementation cost. However, it is not enough just to compress the weight parameters linearly. We also want the result of the compression to be represented in a neural network again for the execution in neuromorphic architectures. In this sense, we \textit{simplify} neural networks. Surprisingly, simplifying neural networks using factorization for the first issue handles the remaining two issues as well. A motivational example is depicted in Fig.~\ref{ex1}. The feedforward network can be considered a 2D filter. Since the 2D kernel is separable, it can be decomposed into a vertical filter and a horizontal filter. Using the decomposition, the single-layer network is factorized into a deeper, sparse network, providing a solution to the second issue. Besides, the weight-shared network can be ``rolled'' to a network with fewer units and connections. This process often require adding delay units for memorizing a previous state of a unit, resulting in a time delay neural network (TDNN), which is a special case of biologically plausible recurrent networks. This TDNN takes input serially and allows us to time-share resources, dealing with the last issue.
Any 2D filter can be decomposed into a set of separable filters using singular value decomposition. We extend this to the bank of 3D filters for typical deep neural networks.
\section{Neural Network Compiler}
In a convolutional neural network for computer vision tasks, the weights in each convolutional layer are typically represented in a 4th-order tensor. The weights in the first fully-connected layer can also be represented in a 4th-order tensor. Let $\chi \in \mathbb{R}^{C \times K_h \times K_w \times F}$ be the weight tensor of a convolutional layer (a fully-connected layer), where $C$ is the number of input channels, $K_h$ and $K_w$ are the height and width of the kernel (the input map), respectively, and $F$ is the number of output feature maps. Applying a tensor decomposition method to the weight tensor, a layer can be factorized into 5 sub-layers. We denote the weight tensors of the sub-layers by $\chi_c, \chi_v, \chi_h, \chi_g$, and $ \chi_i$, respectively, and we represent the weights of each sub-layer in the same format as those of the original layer. The sub-layers are:
\begin{itemize}
\item \textbf{Channel filter layer}: Each unit in this layer is connected to the input units at the same vertical and horizontal position across channels. $\chi_c \in \mathbb{R}^{C \times 1 \times 1 \times R_c}$.
\item \textbf{Vertical filter layer}: Each unit in this layer belongs to a channel and is connected to the inputs at the same vertical position. $\chi_v \in \mathbb{R}^{R_c \times K_h \times 1 \times R_v}$. 
\item \textbf{Horizontal filter layer}: As in the vertical filter layer, each unit acts on a channel and is connected to the inputs at the same horizontal position. $\chi_h \in \mathbb{R}^{R_v \times 1 \times K_w \times R_v}$.
\item \textbf{Code generation layer}: As in the first layer, each unit in this layer acts on the same position across channels. For each position, a code for the original output is generated. $\chi_g \in \mathbb{R}^{R_v \times 1 \times 1 \times R_f}$.
\item \textbf{Inverse transform layer}: As in the first layer, each unit in this layer acts on the same position across channels. For each position, the code is converted into the original output. $\chi_i \in \mathbb{R}^{R_f \times 1 \times 1 \times F}$. 
\end{itemize}
If the original layer is convolutional, the sublayers also become convolutional.  
This factorization makes the network deeper and less dense. We use our own tensor decomposition method, but one can use CP decomposition referring to a recent preprint~\cite{lebedev2014}. In this case, the code generation layer is not needed, and the horizontal filter layer generates the codes directly, and the rank of CP decomposition equals $R_c=R_v=R_f$. After the decomposition, we perform low-rank approximation, greatly simplifying the network. In the case of CP decomposition, a simplified network can be obtained directly. Even for a very high degree of approximation (e.g., 90\% error), the performance of the network is degraded marginally~\cite{lebedev2014}. However, for greater simplification, we fine-tune the factored network after an aggressive approximation, recovering the original performance. 

\begin{figure}
\centering
\begin{subfigure}{.4\textwidth}
  \centering
  \epsfig{figure = 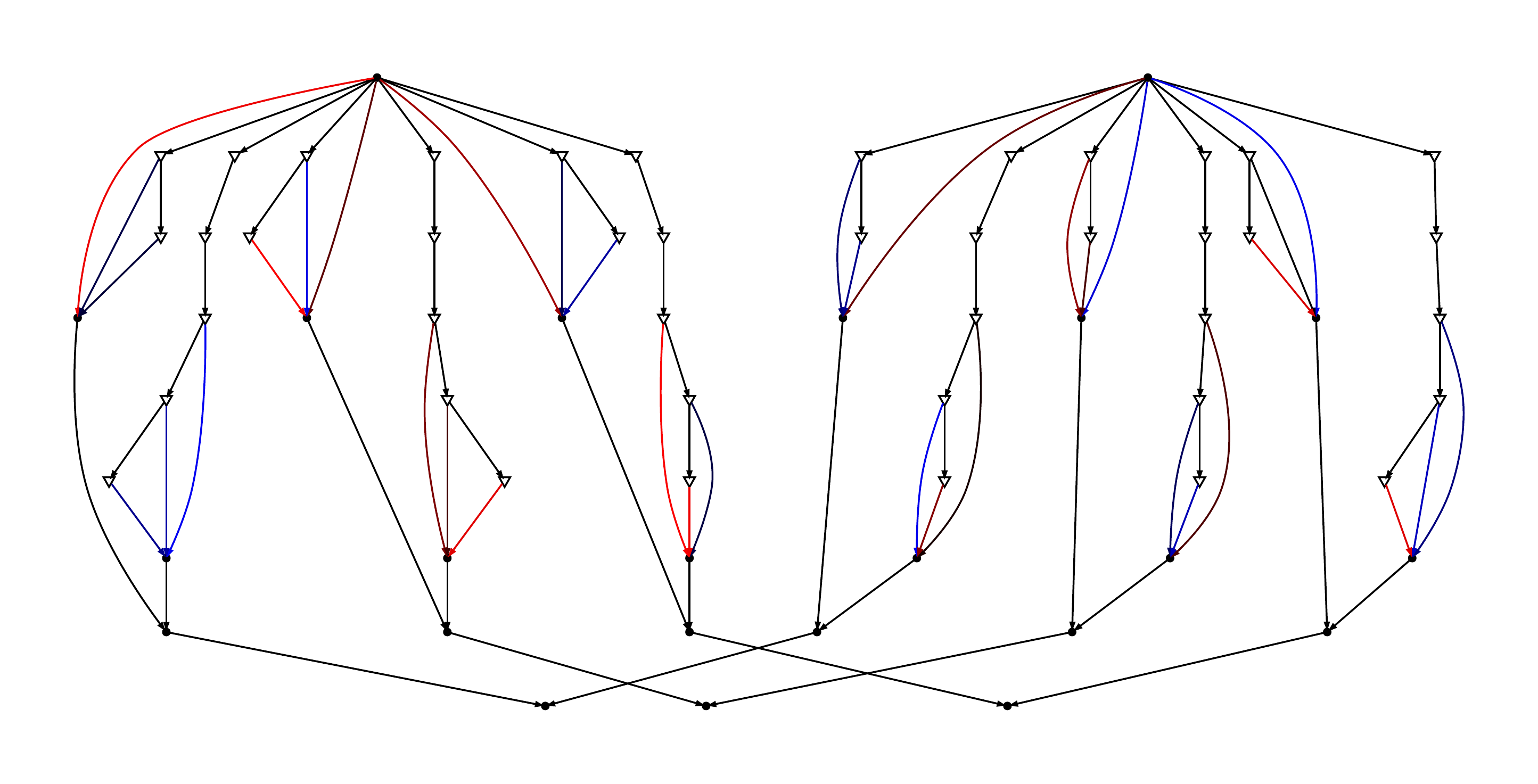, height=1.4in} 
  \caption{0\% synaptic connection reduction with 0\% error}
\end{subfigure}~~~~~~~~~~~~~~~~~~~~~
\begin{subfigure}{.4\textwidth}
  \centering
  \epsfig{figure = 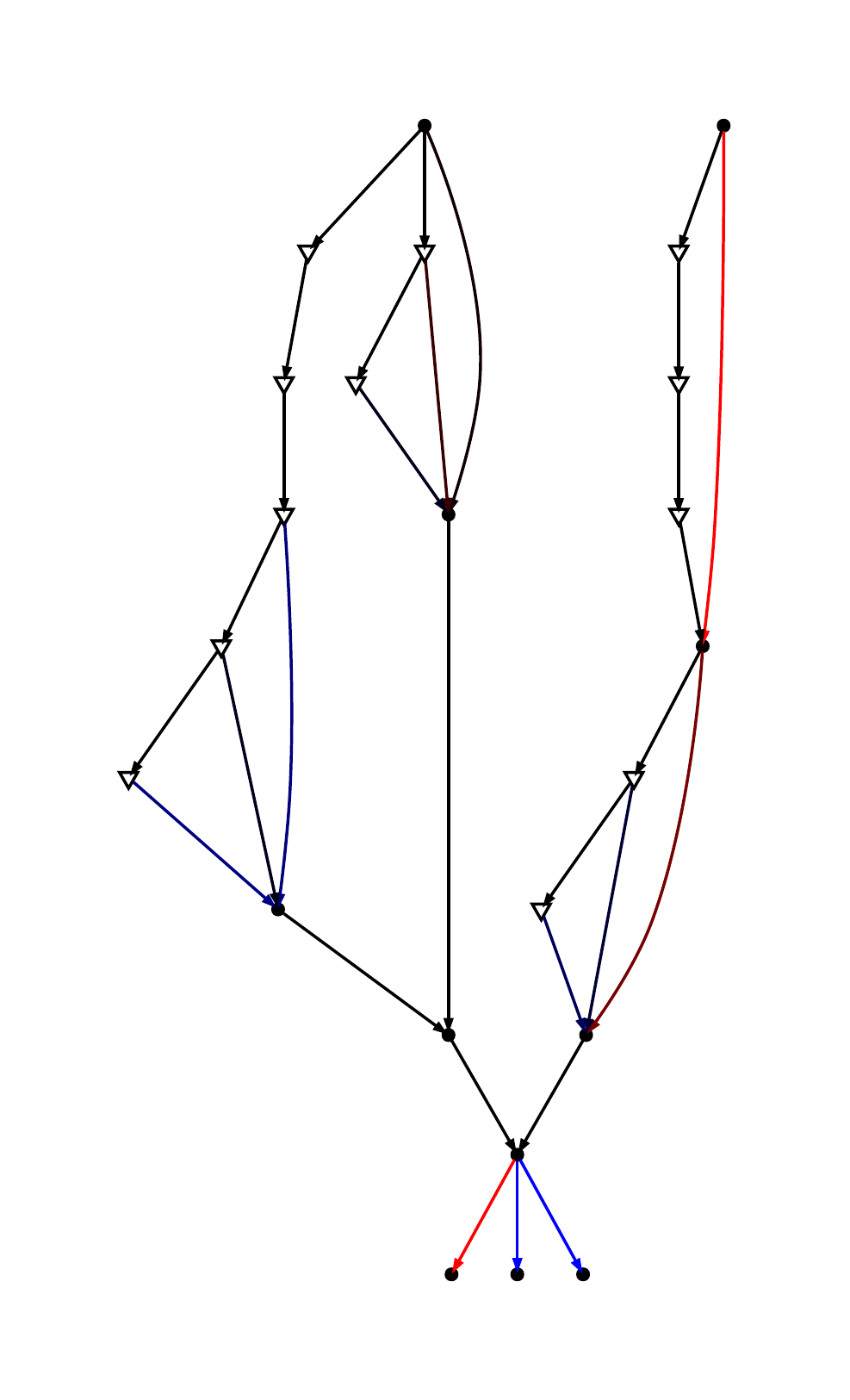, height=1.4in}
  \caption{61.11\% synaptic connection reduction with 50.55\% error}
\end{subfigure}\\
\caption{We randomly generate a single layer network. This network can be considered a feature detector that takes 2 input maps of size $2 \times 3$ and produces 3 feature maps.  The positive-weight connections are colored with red. The negative-weight connections are colored with blue. The darker the color is, the larger the absolute value of the weight is. The connections with weight of 1 are colored with black.}\label{approx}
\end{figure}

%
%
Then, we construct a TDNN corresponding to a factored layer directly. Let $I \in \mathbb{R}^{C\times H \times W}$ be an input signal where $C$ is the number of input maps, and $H$ and $W$ are the height and width of the maps, respectively. 
We first represent the input signal $I$ as a set of $C$ sequences $s_{z}(t)$, $z=0,...,C-1$, such that
\begin{equation}
s_{z}(t) = I(z,y,x),~~t = yW + x,
\end{equation}
for $y=0,...,H-1$, $x=0,...,W-1$. 
We require $W(N-1)$ delay units to obtain $N$ contiguous data points along the vertical axis from an input sequence, and $N-1$ delay units in the case of the horizontal axis.  In the TDNN, the total number of delay units is 
$W(K_h-1)R_c+(K_w-1)R_v$. The delay for the first valid output is $W(K_h-1)+K_w-1$. An example of the TDNNs and its approximation are shown in Figure~\ref{approx}.

\section{Neuromorphic Architecture}
We describe our neuromorphic architecture briefly. Since it is neuromorphic, the architecture may be familiar to readers already. The proposed architecture uses the second-generation neuron model~\cite{vreeken2003spiking} and consists of synapses, neurons, and delay elements, a special type of short-term memories. Synapse and neurons are processing elements. Each synapse stores a real-valued weight and performs the \textit{synaptic operation} that multiplies the incoming input value by the weight. Each neuron produces the sum of the output values of the connected synapses, and it can add a bias and evaluate an activation function such as the sigmoid and rectified linear functions. We call the connections with the weight not equal to one the \textit{synaptic connection}. Unlike the traditional computers, a processing element is not timed-shared for the operations for different synaptic connections; a synapse is dedicated to the operations for a synaptic connection. Thus, for the execution, each synaptic connection is \textit{mapped}, not scheduled, to a synapse, and synaptic connections and synapses have a one-to-one correspondence. Each delay unit is also mapped into a delay element.

\section{Microarchitecture}

\begin{figure}[h]
\centering


  \centering
  \epsfig{figure = 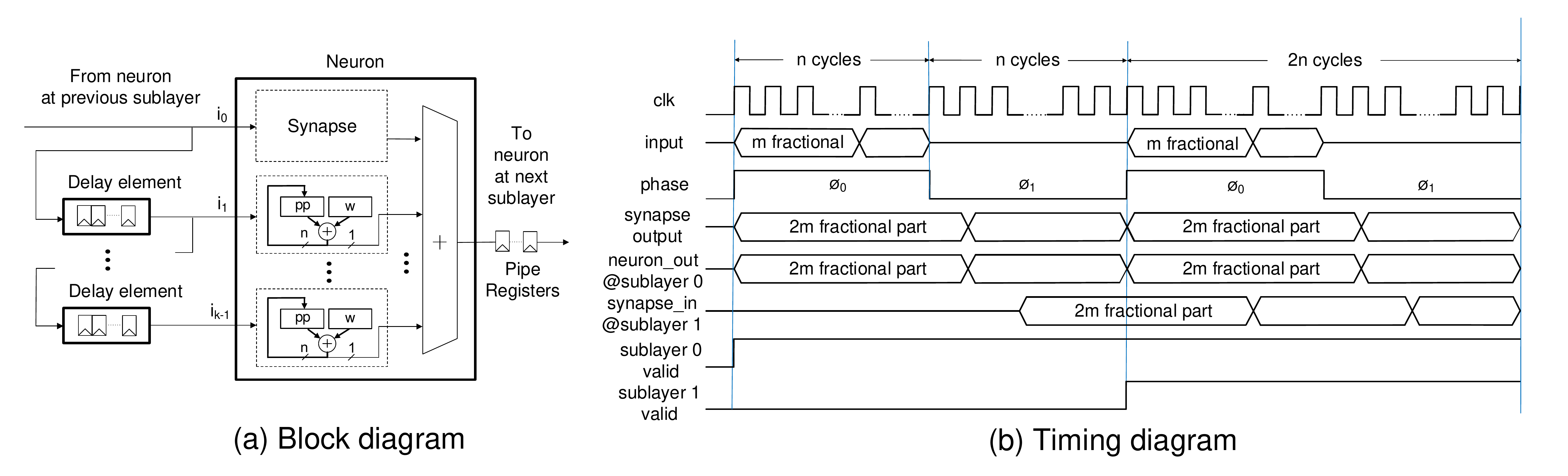, width=5.6in}
\caption{A $k$-input neuron at the horizontal sublayer takes the current input and the last $k-1$ inputs using delay elements. The bias and activation circuits are omitted. The $k$-input adder has the zero latency, and the $2n$-bit result is aligned by the pipeline registers, which truncates the least significant $m$ bits.}\label{timing_diag}
\end{figure}
The proposed architecture can be realized in various ways. We can continue to be bio-inspired to implement a microarchitecture, or can use conventional digital signal processing (DSP) circuits.  At one extreme, one can employ analog spiking neurons to be biologically plausible. At the other extreme, numbers are represented in $n$-bit binary, and conventional logic multiplier, adders, and flip-flops can be used. Especially, the decision on the coding of numbers leads to various trade-offs in circuit area, performance (e.g., classification accuracy), energy-efficiency, fault-tolerance, etc. Various neural codes are introduced in~\cite{gerstner2002spiking}. One can use the rate coding, and a number can be represented as the number of ones in a bitstream, which also enables cost-effective stochastic computing~\cite{alaghi2013survey}. This is an interesting research avenue, but it may require a long bitstream if even 1\% performance degradation is considered critical, and the high robustness provided by the rate coding may not be necessary in digital systems. We take a moderate approach, and use the conventional binary coding. However, we employ the fixed-point, bit-serial representation, which provides the following strengths. 1) Unlike the rate coding, the precision increases exponentially in the number of bits, so we can easily avoid the performance degradation due to the quantization error. If the performance degradation is permitted, it saves more implementation cost to perform a higher degree of low-rank approximation to the network. 2) We can use the cost-effective, bit-serial arithmetic~\cite{denyer1985vlsi}. Only a handful of logic gates and sequential elements are needed to implement bit-serial adders and bit-serial multipliers. 3) We can leverage the existing integrated circuit design tools fully because they understand the binary coding well.

In our implementation, a number is represented as an $n$-bit stream where the least significant bit (LSB) comes first. To represent signed real numbers in the $n$-bit, we use the fixed-point format and let $m$ denote the fractional bit-width. Figure~\ref{timing_diag}(a) depicts a $k$-input neuron.
 The $k$-input neuron comprises $k$ synapses, a $k$-input bit-serial adder to add up the outputs of the $k$ synapses, a bias circuit and an activation circuit to evaluate an activation function. A synapse is implemented by an $n$-bit register to store the weight and a bit-serial multiplier, which mainly consists of $n$ full adders and several registers to store intermediate results. The output of the $k$-input adder can go to the bias circuit followed by the activation circuit, or directly go to the output of the neuron. A delay element is realized by a 1-bit $n$ stage shift register. Figure~\ref{timing_diag}(b) depicts a timing diagram of our implementation. It takes $2n$ cycles to compute the product of two $n$-bit numbers. To put the product back to the $n$-bit format, we truncate the least significant $m$-bits, and the first valid output becomes available $m$ cycles later. Thus, the synapse comes to have an inherent $m$ cycle delay. To simplify the design, we pad a $2n-m$ cycle delay at the output of the neuron, and the registers for this delay can be moved forward to improve timing (i.e., they are used as pipeline registers and can be re-timed). Then, the outputs of a sublayer become available $2n$ cycles after the inputs arrive, and all the synapses in the system start a synaptic operation at the same time at every $2n$ cycle. Delay elements perform shift only for the first half of the $2n$ cycle period. 

\section{Experiments}
We implement the neural network compiler using pylearn2~\cite{pylearn2_arxiv_2013}. We use our own tensor approximation method, but CP-decomposition can also perform similarly. Neural network models are trained using Nvidia GTX780 GPU and are fed into the compiler. We chose Artix7 XC7A100TCSG324-1 field-programmable gate array (FPGA) as the target platform. Note that this FPGA has relatively small capacity and  40 times larger FPGAs are available in the market. Our experiments can be considered in two ways. First, FPGAs are considered a general-purpose, not-highly-optimized neuromorphic processor, and the experiments are regarded as making software for the processor in part using existing hardware synthesis tools. FPGAs already have plastic connections, and processing elements and memories are mixed in space. The SRAM-based look-up tables (LUTs) serve as memories as well as processing elements. Second, our experiments can also be considered to build a prototype of an application-specific neuromorphic processor. While the neural network complier generates a hardware model in verilog in an application-specific fashion, we can easily extend it into a general-purpose hardware model if programmable interconnects are available. We elaborate the generated hardware models for the target platform using Xilinx Vivado. The shift registers in delay elements are refined into an SRAM-based shift register, not a chain of flip-flops, and even long shift registers are implemented efficiently.

\begin{table}\scriptsize
\centering
\caption{The resulting TDNNs and their implementations for varying degrees of approximation.}\label{compare}
\begin{tabular}{|c||c|c|c|c|c|c|c|c|c|}
\hline
Approx. &\multicolumn{4}{c}{Classification Accuracy}&\multicolumn{2}{|c|}{TDNN}&\multicolumn{3}{c|}{Implementation}\\\cline{2-10}
 error	& Original  & Post-approx.  & Post-tuning  & Post-imp.  & Params & Delay units & Total LUTs & Mem. LUTs  & Power (W)\\ \hline 
10\%   & 0.9234   & 0.9131   & 0.9226   & 0.9155   & 3226   & 8316   & NA   & NA   & NA      \\
16\%  & 0.9234   & 0.908   & 0.9217   & 0.9146   & 2274   & 7857   & 53467   & 4189   & 2.548      \\
31\%   & 0.9234   & 0.8287   & 0.9154   & 0.9109   & 1200   & 6588   & 31223   & 4065   & 1.483    \\
40\%   & 0.9234   & 0.7684   & 0.9055   & 0.9008   & 854   & 5670   & 23810   & 2853   & 1.222     \\
51\%   & 0.9234   & 0.5403   & 0.8943   & 0.8943   & 564   & 4779   & 16954   & 3007   & 0.869     \\
 61\%  & 0.9234   & 0.5135   & 0.8377   & 0.8368   & 376   & 3186   & 11343   & 1581   & 0.669      \\
 82\%  & 0.9234   & 0.3586   & 0.6115   & 0.5989   & 132   & 1566   & 4754   & 779   & 0.363    \\
(ConvNet2)   & 0.9913  & 0.1010  & 0.9764   & 0.9764  & 1288   & 8844   & 45843   & 6809   & 1.120      \\
\hline 
\end{tabular}
\end{table}
To demonstrate our approach, we use a softmax regression for hand-written digit classification. The neural network for the regression does not require the hardware for activation functions and pooling, which allows us to focus on the reduction of the implementation cost by the low-rank approximation.  The original neural network has 7840 weight parameters. We train the network using the MNIST data set and achieve 92.34\% accuracy. For various degrees of approximation, the neural network compiler generates TDNNs and the corresponding hardware models. Using Vivado, the hardware model is synthesized, and the power consumption and the hardware cost in terms of LUTs are obtained. All the designs operate at 160MHz. Table~\ref{compare} compares the results. The classification accuracy is evaluated after the approximation, fine-tuning, and implementation. The post-implementation accuracy is the performance of the actual system and reflects the finite-precision effects. We use 7 bits for the fractional part. For TDNNs, the number of parameters and the number of delay units indicate the size of the models. The number of parameters equals the number of synapses, and LUTs in the FPGA are mainly used to implement synapses. A large number of delay units are realized by a small number of memory LUTs, and a memory LUT occupies a similar space to a LUT in the chip. Thus, the hardware cost is mainly determined by the number of synapses (parameters). 
 The original feedforward network with 7840 parameters cannot be implemented in the small FPGA in a neuromorphic fashion, but the TDNN conversion and the approximation make it possible. The TDNN with 3226 parameters is slightly larger to be fitted to the FPGA, but the smaller TDNNs are easily contained in the target system.    In neuromorphic architectures, the reduction of parameters leads to the reduction of synapses, which is directly translated into the reduction of hardware cost and power/energy consumption. In this experiment, the parameter reduction rate with negligible accuracy loss is not that high possibly because it is a single-layer network. For recent deep architectures, more than 100$\times$ parameter reduction is possible with marginal loss of accuracy~\cite{lebedev2014}.  

\begin{table}\scriptsize
\centering
\caption{The results of the neural network compiler for a convolutional neural network.}\label{layer}
\begin{tabular}{|c||c|c|c|c|c|c|c|c|}
\hline
         & Original   &  \multicolumn{3}{|c|}{ConvNet1} &  \multicolumn{3}{|c|}{ ConvNet2} \\\cline{3-8}
     	& Params     & Approx. error & Params & Param. reduction   & Approx. error & Params & Param. reduction  \\ \hline 
Layer 1  &   1600   & 12\%  & 588  & 2.72$\times$ & 50\%   & 168   & 9.52$\times$       \\
Layer 2  &   102400   &  88\%     & 2474   &  41.39$\times$  & 94\%   & 829   & 123.52$\times$      \\
Layer 3  &  5760     &  40\%     &  1786   &  3.23$\times$  & 83\%   & 291   & 19.79$\times$       \\
\hline
\end{tabular}
\end{table}
\begin{figure}
	  \centering
\begin{subfigure}{.32\textwidth}
  \centering
  \epsfig{figure = 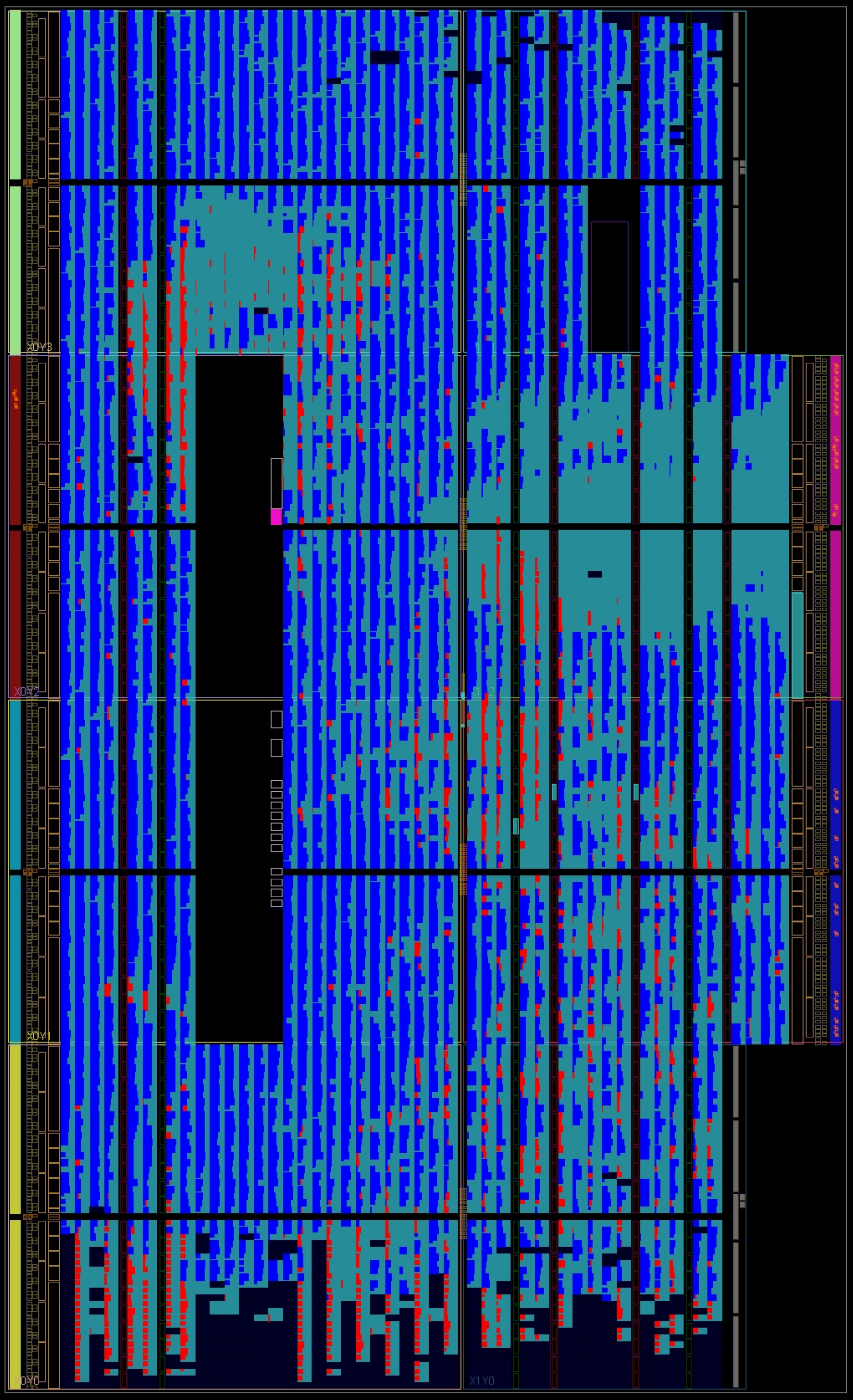,  width = 1.7in}
  \caption{}
\end{subfigure}
\begin{subfigure}{.32\textwidth}
  \centering
  \epsfig{figure = 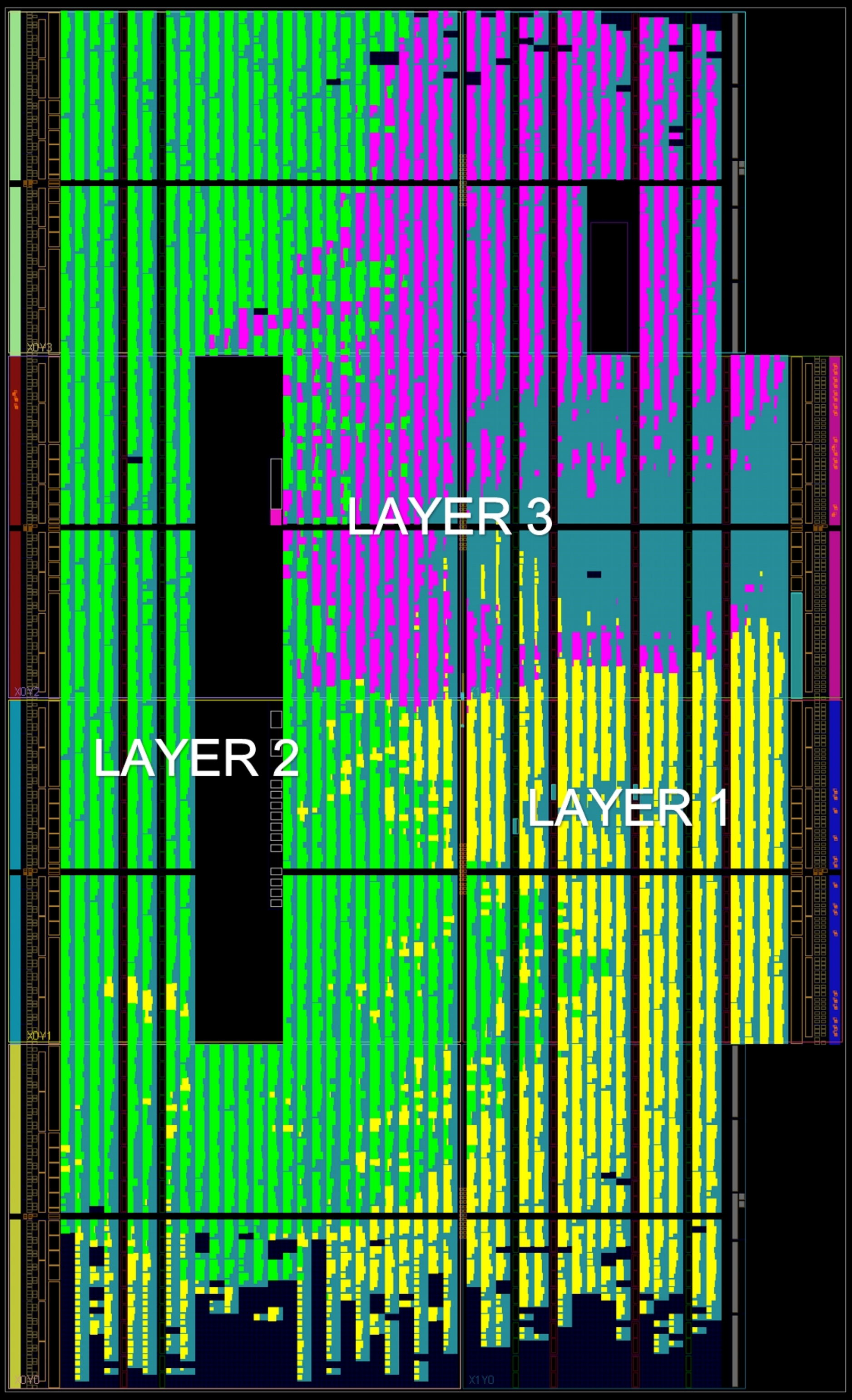,  width = 1.7in}
  \caption{}
\end{subfigure}
\begin{subfigure}{.32\textwidth}
  \centering
  \epsfig{figure = 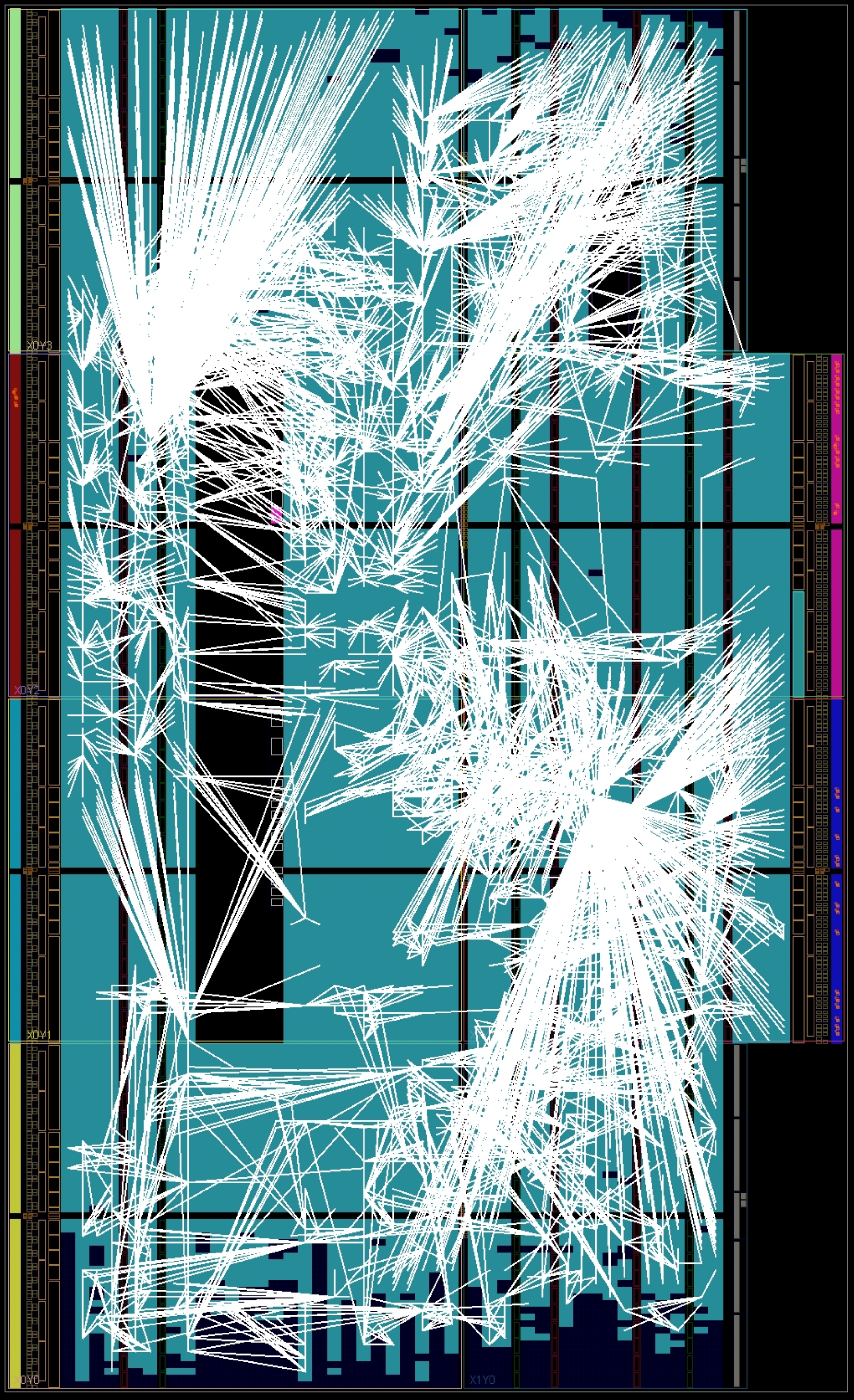, width = 1.7in}
  \caption{}
\end{subfigure}
 \caption{\textbf{Neural network on a chip}: (a) The neurons and the delay elements in ConvNet2 are highlighted in blue and red, respectively. (b) The three layers of ConvNet2 are highlighted in yellow, green, and magenta, respectively. (c) The connectivity of ConvNet2 is shown. }\label{nnoc}
 \end{figure}
To show that our approach can be extended to multi-layer neural networks, we use a convolutional neural network (CNN) for the same task. The network has 2 convolutional layers with max-pooling and the rectified linear function as the activation function, and a final softmax layer. The baseline model has 109,760 parameters and achieves 99.13\% classification accuracy. From the baseline, the compiler creates two approximated networks with reduced parameters. The first network (ConvNet1) maintains the classification accuracy after re-training, and the second network (ConvNet2) is simplified until the resulting hardware fits to our target platform. The results are summarized in Table~\ref{compare} and~\ref{layer}.
ConvNet2 is simplified too much to be fitted to the small FPGA sacrificing some accuracy, but ConvNet1 will provide the original accuracy on an industrial FPGA. Our neuromorphic implementation of ConvNet2 provides the exactly same accuracy as the model satisfying our design goal that is not to sacrifice the accuracy except the low-rank approximation. 
For this task, to our best knowledge, the best accuracy achieved on neuromorphic architectures is 91.94\%~\cite{esser2013cognitive}.

\begin{figure}[h]
\centering
\epsfig{figure = 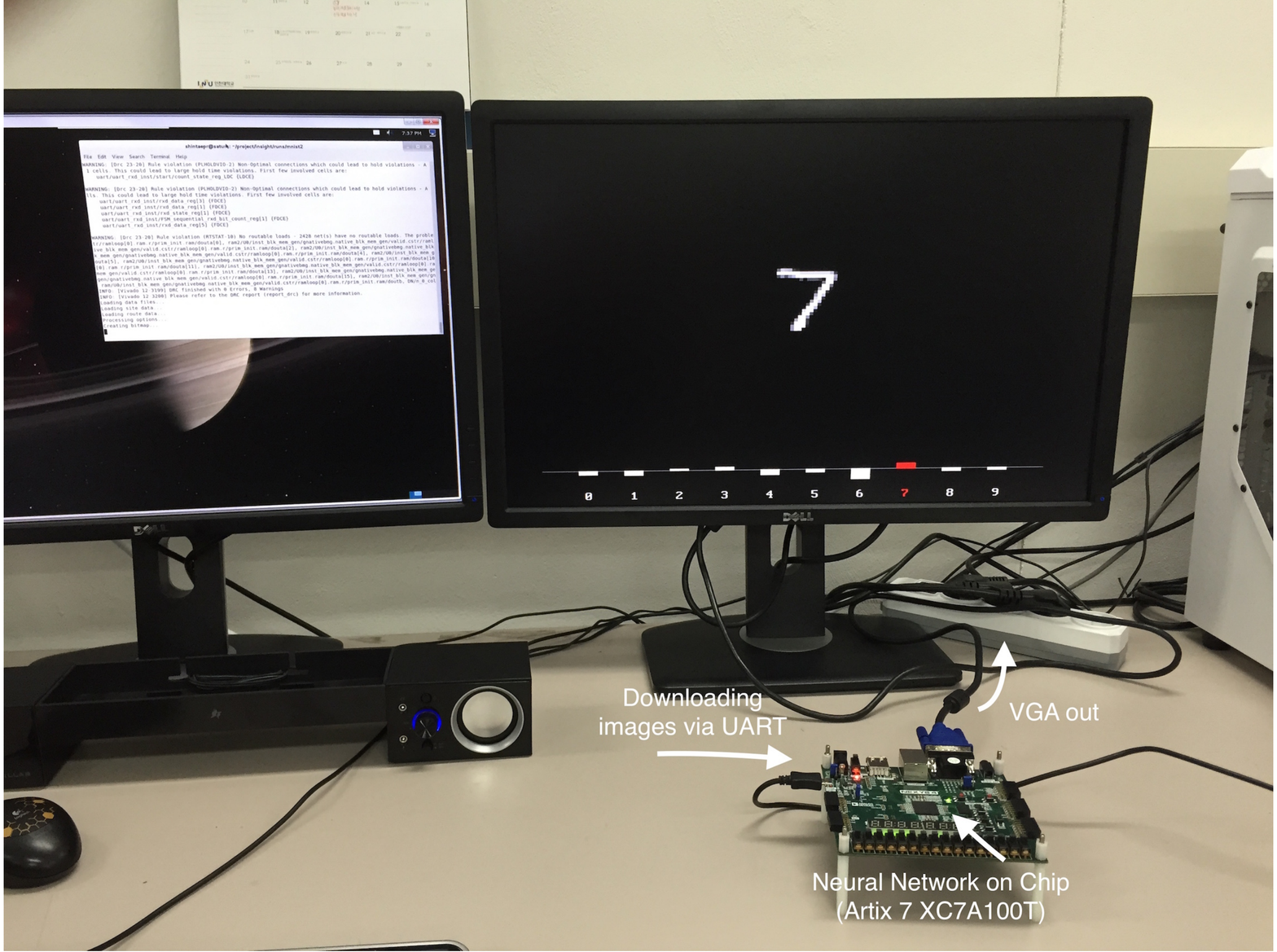, width = 3.5in}
\caption{Our neuromorphic system based on a FPGA is executing a neural network originally with 100K weight parameters for the hand-written digit classification task. Now both software and hardware are bio-inspired. The classification accuracy of the system is 97.64\% even though a small FPGA is used, and it processes an input image in 156.8$\mu$s. The internal block RAMs are used as the frame buffer, and any external memory is not used in this system, which shows that it is a non-Von Neumann architecture.}\label{demo}
\end{figure}
A TDNN model is refined to a network of silicon neurons and delay elements preserving the topology, and the IC design tool places the neurons and the delay elements in a space and connects them automatically. This provides interesting visualization of neural networks. ConvNet2 on the FPGA is shown in Fig.~\ref{nnoc}. The network on the FPGA is stimulated by images in the test set of MNIST, which are transferred via UART, and the output of the network is visualized in a screen as shown in Figure~\ref{demo}. The neuromorphic system successfully classifies images in real-time, and the results match with the simulation exactly. 

\section{Discussion}
In this paper we have presented a neuromorphic computing system that is newly designed from the microarchitecture to the compiler in order to forward-execute neural networks with minimum energy consumption.  We have also demonstrated that the parameter reduction via low-rank approximation is much more valuable than considered in the community. This neuromorphic system can scale by simply using a larger FPGA or by fabricating a custom chip. 
Since more than 40 times larger FPGAs than the target platform are available in the market, with our approach, it now becomes easy to build neuromorphic computing system that can execute neural networks with more than 4 million real-valued parameters, fully leveraging existing integrated circuit design techniques.  Our neuromorphic system has just built and is not fully optimized. Through all the levels of the computing stack, there is huge room for optimization. Our future work may include the circuit-level optimization and neural network transformations to minimize the implementation cost. 
 

%
%

\bibliographystyle{unsrt}
\bibliography{scibib}

%
%
%
%

\end{document}